\documentclass[11pt,a4paper]{article}
\usepackage[hyperref]{acl2020}
\usepackage{times}
\usepackage{latexsym}
\usepackage{amsmath}
\usepackage{multirow}
\usepackage{graphicx}
\usepackage{tabularx}

\usepackage{microtype}

\aclfinalcopy 

\title{Spotify at TREC 2020: \\Genre-Aware Abstractive Podcast Summarization}

\author{Rezvaneh Rezapour\thanks{$^{*}$\ Work done while at Spotify} \\ 
University of Illinois at Urbana-Champaign\\ Urbana, IL, USA\\
\texttt{rezapou2@illinois.edu}\\
\And
Sravana Reddy\\
Spotify\\Boston, MA, USA\\
\texttt{sravana@spotify.com}\\
\AND
Ann Clifton\\
Spotify\\New York, NY, USA\\
\texttt{aclifton@spotify.com}\\
\And
Rosie Jones\\
Spotify\\Boston, MA, USA\\
\texttt{rjones@spotify.com}

}

\date{}

\begin{document}
\maketitle
\begin{abstract}
This paper contains the description of our submissions to the summarization task of the Podcast Track in TREC (the Text REtrieval Conference) 2020. The goal of this challenge was to generate short, informative summaries that contain the key information present in a podcast episode using automatically generated transcripts of the podcast audio. Since podcasts vary with respect to their genre, topic, and granularity of information, we propose two  summarization models that explicitly take genre and named entities into consideration in order to generate summaries appropriate to the style of the podcasts. Our models are abstractive, and supervised using creator-provided descriptions as ground truth summaries. The results of the submitted summaries show that our best model achieves an aggregate quality score of 1.58 in comparison to the creator descriptions and a baseline abstractive system which both score 1.49 (an improvement of 9\%) as assessed by human evaluators.
\end{abstract}

\section{Introduction}
Advancements in state-of-the-art sequence to sequence models and transformer architectures \citep{vaswani2017attention,dai2019transformer,nallapati-etal-2016-abstractive,ott2019fairseq,lewis-etal-2020-bart}, and the availability of large-scale datasets have led to rapid progress in generating near-human-quality coherent summaries with abstractive summarization. The majority of effort in this domain has been focused on summarizing news datasets such as CNN/Daily Mail \citep{hermann2015teaching} and XSum \citep{narayan-etal-2018-dont}. 
Podcasts are much more diverse than news articles in the level of information as well as their structure, which varies by theme, genre, level of formality, and target audience, motivating the TREC 2020 podcast summarization task \cite{trec2020podcastnotebook}, where the objective is to design models to produce coherent text summaries of the main information in podcast episodes.

While no ground truth summaries were provided for the TREC 2020 task, the episode descriptions written by the podcast creators serve as proxies for summaries, and are used for training our supervised models. We make some observations about the creator descriptions towards understanding what goes into a `good' podcast summary. Our first observation is that the descriptions vary stylistically by genre.
For instance, the descriptions of sports podcasts tend to list a series of small events and interviews, usually anchored on the 
names of players, coaches, and teams. In contrast, descriptions of true crime podcasts tend to contain suspenseful hooks, and avoid giving away much of the plot (Table \ref{tab:genre_examples}). Some podcasts include the names of hosts or guests in their descriptions, while others only talk about the main theme of the episode. 

\begin{table}[]
\footnotesize
  \centering
  \resizebox{\linewidth}{!}{
  \begin{tabular}{|c|p{2in}|}
  \hline
     \textbf{Sports} & In EP 4 \textit{Jon Rothstein} is joined by Arizona State Head Coach \textit{Bobby Hurley}, Creighton Guard \textit{Mitchell Ballock}, and Merrimack Head Coach \textit{Joey Gallo} checks in on the Hustle Mania Hotline. This is \textit{March} and \textit{Jon} has a special message for all of you heading into the NCAA tournament.\\
  \hline
    \textbf{True Crime} & Sometimes, \textit{it takes years to connect a killers crimes}. On March 6th 1959 \textit{a man} was born who would, eventually, be tried for the murder of a single woman. I\textit{t would take years for police to connect him to 4 other murders}. Years and a clever investigator who got the DNA he needed. \\
  \hline
    \textbf{Comedy} & This weeks episode is a little bit of a throwback.. We have \textit{one of our close high school friends Zak on the show}, and he is just a \textit{blast}! We discuss relationships, our favourite movies and \textit{our experience's with high school}! This episode is pretty chill, so you better have a drink for this one!\\
  \hline
  \end{tabular}}
  \caption{Examples of creator descriptions for a different genres of podcast episodes, highlighting the prevalence of named entities in sports, suspense in true crime, and colloquial language in comedy.}
  \label{tab:genre_examples}
\end{table}

By definition, a good summary should be concise, preserve the most salient information, and be factually correct and faithful to the given document \cite{nenkova2011automatic,maynez-etal-2020-faithfulness}. In addition, we believe that since users' expectations of summaries are shaped by the creator descriptions and the style and tone of the content, it is important to generate summaries that are appropriate to the style of the podcast.

In this work, we present two types of summarization models that take these observations into consideration. 
The goal of the first model, which we shorthand as `category-aware', is to generate summaries whose style is appropriate for the genre or category of the specific podcast. The second model aims to maximize the presence of named entities in the summaries.


\section{Data Filtering}
The Spotify Podcast Dataset consists of 105,360 episodes with transcripts and creator descriptions \cite{clifton-etal-2020-100000}, and is provided as a training dataset for the summarization task. We applied a set of heuristic filters on this data with respect to the creator descriptions.

\begin{itemize}
  \item Removed episodes with unusually short (fewer than 10 characters) or long (more than 1300 characters) creator descriptions.
  \item Removed episodes whose descriptions were highly similar to the descriptions of other episodes in the same show or to the show description. Similarity was measured by taking the TF-IDF representation of a description and comparing it to others using cosine similarity.
  \item Removed ads and promotional sentences in the descriptions. We annotated a set of 1000 episodes for such material, and trained a classifier to detect these sentences; more details are described in \newcite{podcastaddetect}. 
\end{itemize}

The dataset after these filters consists of $90055$ episodes. We held out $1000$ random episodes for test and validation sets for development, and used $88055$ episodes for training our models. Our final submissions were made on the task's test set of $1027$ episodes  disjoint from the training data.

\section{Abstractive Model Framework}
Our summarization models are built using BART \citep{lewis-etal-2020-bart}, which is a denoising autoencoder for pretraining sequence to sequence models. To generate summaries, we started from a model pretrained on the CNN/Daily Mail news summarization dataset\footnote{https://huggingface.co/facebook/bart-large-cnn} and then fine-tuned it on our podcast transcript dataset with respect to the two proposed models (described below in \S\ref{sec:models}). 

We used the implementation of BART in the Huggingface Transformers library \citep{wolf-etal-2020-transformers}  with the default hyperparameters. We set the batch size to 1, and made use of 4 GPUs for training. The best model with the highest {\sc ROUGE-L} score on the validation set was saved. The maximum sequence limit we set for the input podcast transcript was 1024 tokens (that is, any material beyond the first 1024 tokens was ignored).
We constrained the model to generate summaries with at least 50 and at most 250 tokens. 

This setup is the same as the {\em bartpodcasts} model described in the task overview  \citep{trec2020podcastnotebook}, and is one of our baselines for comparison. We also compare our models to the {\em bartcnn} model baseline, which is the out of the box BART model trained on the CNN/Daily Mail corpus.

\begin{figure*}[ht]
 \centering
 \includegraphics[width=6in]{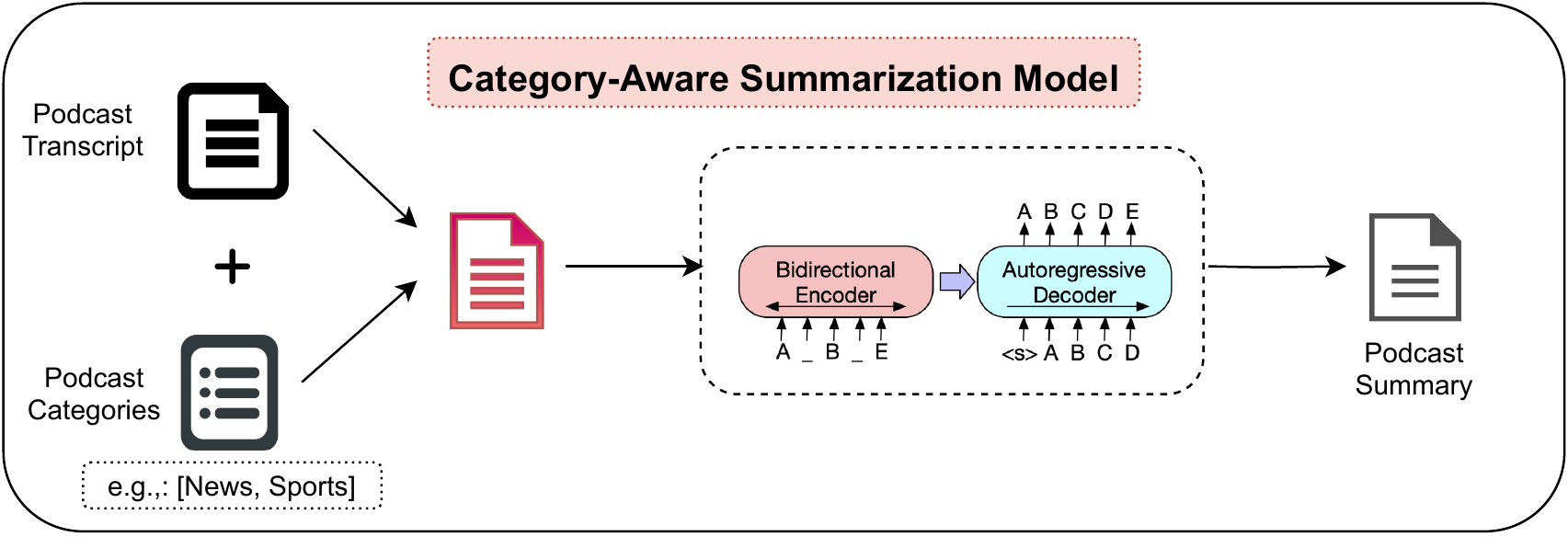}
 \caption{Sketch of the Category Aware Summarization Model}
 \label{fig:category_aware_model}
\end{figure*}

\section{Models} \label{sec:models}

\subsection{Category-Aware Summarization}
The idea of building a category-aware summarization model is motivated by the hypothesis that selecting what is important to a summary depends on the topical category of the podcast. 
This hypothesis is based upon observations of the creator descriptions in the dataset. For example, descriptions of `Sports' podcast episodes tend to contain names of players and matches, `True Crime' podcasts incorporate a hook and suspense, and `Comedy' podcasts are stylistically informal (Table \ref{tab:genre_examples}). 

While we expect some of these patterns to be naturally learned by a supervised model trained on the large corpus, we nudged the model to generate stylistically appropriate summaries by explicitly conditioning the summaries on the podcast genre. 
Some previous work approaches this problem by concatenating a vector corresponding to an embedding of the conditioning parameters to the inputs in an RNN \cite{ficler-goldberg-2017-controlling}. More recent work simply concatenates a token corresponding to the parameter to the text input in sequence-to-sequence transformer models \citep{2020t5}. 
We experimented with a few ways of encoding the category labels in the summarization model, and settled upon prepending the labels as special tokens to the transcript as the input during training and inference, as shown in Figure \ref{fig:category_aware_model}. For the episodes with multiple category labels, we concatenated the labels and included them all as distinct special tokens, concatenated in a fixed (lexicographic) order. 

\begin{table}[ht!]
\small
  \centering
  \begin{tabular}{|l|l|}\hline
  \textbf{Categories} & \textbf{Counts}\\
\hline
Arts & 81 \\
Business & 97\\
Comedy & 136\\
Education & 118 \\
Fiction & 5 \\
Games & 7 \\
Government & 3\\
Health & 94 \\
History & 16 \\
Kids \& Family & 28\\
Leisure & 45 \\
Lifestyle \& Health & 42 \\
Music & 33 \\
News & 15 \\
Religion \& Spirituality & 92\\
Science & 11 \\
Society \& Culture & 96\\ 
Sports & 137\\
Stories & 37\\
TV \& Film & 41 \\
Technology & 17 \\
 True Crime & 28 \\
\hline
  \end{tabular}
  \caption{Set of collapsed category labels and their distribution on the task test set. Some episodes are associated with multiple labels.}
  \label{tab:categories}
\end{table}

The podcast categories are not given in the metadata files distributed with the podcast dataset. However, they can be derived from the RSS headers associated with each podcast under the {\em itunes:category} field. In our case, we leveraged the labels assigned in the Spotify catalog.
The category labels in the catalog are mostly the same as the labels in the RSS header, with some minor changes. 
The taxonomy of iTunes categories changes over time, leading to different labels for semantically similar categories. We chose to 
heuristically collapse similar categories together, such as `Sports' with `Sports \& Recreation', giving a set of 22 categories (Table \ref{tab:categories}). 

After prepending the labels as special tokens to the transcript, we fine-tuned the BART baseline. Since training BART was computationally expensive, we only trained the model for a maximum of 2 epochs and generated abstractive summaries using 1 ({\em category-aware-1epoch}) and 2 ({\em category-aware-2epochs}) epochs, separately.

\subsection{Named Entity Biased Model}
Named entities are information-dense, and many podcast topics tend to center around people and places, making them appropriate for inclusion in summaries. We furthermore (1) observed that named entity mentions are prevalent in creator descriptions, and (2) surveyed a group of podcast listeners about what they look for in a summary, where the presence of names and locations was highlighted. Therefore, our second summarization algorithm aims to bias the model to maximize named entities in summaries. 

Our model takes a two-step approach by extracting a portion of the transcript that is both highly relevant to the episode and tends to contain salient named entities. The extracted portion of the transcripts is the input for both training and inference. This approach is similar to the previous approaches that implement an extractive summarizer followed by an abstractive model \citep{ijcai2017-574}. 

Our proposed extractive summarizer is similar to TextRank \cite{mihalcea2004textrank} and consists of the following steps:

\begin{itemize}
  \item We first identified named entities for the entire transcript of each episode by using a BERT token classification model trained\footnote{https://huggingface.co/dbmdz/bert-large-cased-finetuned-conll03-english} on the CoNLL-03 NER data.
  \item We divided each episode into segments corresponding to about 60 seconds of audio.
  \item Every segment was represented in its original form $s$, and as a list of its named entity tokens $t$. All pairwise similarities between segments were computed as the proportion of word overlap between them. We computed both $Sim_s(i, j)$, the similarity between the original forms of the segments $i$ and $j$, and $Sim_t(i, j)$, the similarity between the list of named entity tokens in the segments.
  \item Degree centralities for each segment $C_s(i) = \sum_j Sim_s(i, j)$ and $C_t(i) = \sum_j Sim_t(i, j)$ were computed. The top 7 segments by $C_s(i)$ and the top 3 non-overlapping segments by $C_t(i)$ were extracted and presented in position order as the extractive summary. We chose this heuristic to encourage the extracted segments to be semantically relevant and contain named entities.
\end{itemize}

The BART baseline was then fine-tuned for 3 epochs on extracted segments as input (rather than the first 1024 tokens of the transcript). We name this the {\em coarse2fine} model. 

\begin{table}[ht]
\small
\centering
\resizebox{\linewidth}{!}{
\begin{tabular}{|l|c|c|c|}
\hline
    & \multicolumn{3}{c|}{\textbf{ROUGE-L}}\\ \cline{2-4} 
    & \textbf{Precision} & \textbf{Recall} & \textbf{F1} \\ \hline
{\em bartcnn} & 8.49& \textbf{27.19 } & 11.3\\ \hline
{\em bartpodcasts} & 20.78& 21.01& 16.64\\ \hline
\textbf{\em category-aware-1epoch}& 22.70& 20.82& 17.62\\ \hline
\textbf{\em category-aware-2epochs}& \textbf{25.75}& 19.86& \textbf{18.42}\\ \hline
\textbf{\em coarse2fine} & 15.76 & 18.75& 13.59\\ \hline
\end{tabular}}
\caption{{\sc ROUGE} scores against all of the creator descriptions in the test set.}
\label{tab:rougeall}
\end{table}
\section{Results}
The TREC task uses two evaluation methods to analyze  performance: (1) {\sc ROUGE-L} scores to compare the generated summaries against the creator descriptions as the reference, and (2) human evaluations facilitated by the National Institute of Standards and Technology (NIST) to asses the quality of the generated summaries with respect to the episode content. 

\begin{table}[]
\small
\centering
\resizebox{\linewidth}{!}{
\begin{tabular}{|l|c|c|c|}
\hline
    & \multicolumn{3}{c|}{\textbf{ROUGE-L}}\\ \cline{2-4} 
    & \textbf{Precision} & \textbf{Recall} & \textbf{F1} \\ \hline
{\em bartcnn} & 10.87 & \textbf{29.85} & 14.64\\ \hline
{\em bartpodcasts} & 27.89 & 25.51 & \textbf{22.15}\\ \hline
\textbf{\em category-aware-1epoch}& 27.49 & 22.45 & 20.54\\ \hline
\textbf{\em category-aware-2epochs}& \textbf{30.63} & 23.23 & 22.02\\ \hline
\textbf{\em coarse2fine} & 22.85 & 20.57 & 17.12\\ \hline
\end{tabular}}
\caption{{\sc ROUGE} scores aggregated over only those $71$ creator descriptions that were rated good or excellent.}
\label{tab:rougeeg}
\end{table}

\begin{table}[]
\small
\centering
\resizebox{\linewidth}{!}{
\begin{tabular}{|l|c|c|c|c|c|}
\hline
    & \textbf{E} & \textbf{G} & \textbf{F} &\textbf{B} & {\bf score}\\ \hline
Creator Description & 15.64 & 24.02 & 34.08 & 26.26 & 1.45\\\hline
Cleaned Creator Description & 18.44 & 21.23 & 32.96 & 27.37 & 1.49 \\\hline
{\em bartcnn} & 5.59 & 13.97 & 49.26 & 31.28 & 0.99\\ \hline
{\em bartpodcasts} & 13.97 & 27.93 & 37.43 & 20.67 & 1.49\\ \hline
\textbf{\em category-aware-1epoch} & 14.53 & 27.37 & 38.55 & 19.55 & 1.51 \\ \hline
\textbf{\em category-aware-2epochs} & 17.88 & 21.79 & 43.02 & 17.32 & 1.58 \\ \hline
\textbf{\em coarse2fine} & 10.06 & 21.79 & 29.61 & 38.55 & 1.30 \\ \hline
\end{tabular}}
\caption{Distribution of EGFB annotations over 179 judged episodes (percentages), and the aggregate quality score, computed as a weighted average with E=4, G=2, F=1, and B=0.}
\label{tab:egfb}
\end{table}

Table \ref{tab:rougeall} shows the results of our models against the {\em bartcnn} and {\em bartpodcasts} baselines on the whole test set. 
Compared to the baselines, both {\em category-aware} models achieved higher {\sc ROUGE} precision and F1-scores. While {\em bartcnn} has the highest recall ($27.19\%$), our {\em category-aware-2epochs} summarizer outperformed other models achieving $27.75\%$ precision and $18.42\%$ F1-score. 


Since creator descriptions are of varying quality, we performed another evaluation, selecting only those episodes whose creator descriptions were labeled as `Good' or `Excellent' by the evaluators (Table \ref{tab:rougeeg}). The {\em category-aware-2epochs} model still achieves the highest {\sc ROUGE} precision, and has comparable F1 with the {\em bartpodcasts} baseline.

High precision across our models shows that prepending the categories to the transcripts resulted in more {\em relevant} information in the generated summaries, with tradeoffs against recall.
Tables \ref{tab:rougeall} and \ref{tab:rougeeg} show that the {\em coarse2fine} model only improved over {\em bartcnn}. We observe that the summaries do consist of more named entities compared to the other models (Table \ref{tab:anecdotes}) as intended. We observe that the discontinuities in the extracted segments result in more incoherent abstractive summaries than the models that consume a contiguous block of text, and sometimes contain `hallucinated' content. 

Looking into the manual assessment of the summaries (Table \ref{tab:egfb}), we see that the {\em category-aware-2epochs} model generated more summaries labeled as `Excellent' and fewer summaries labeled as `Bad' compared to the other models and is comparable in quality to the creator descriptions. 


\begin{table*}[ht]
\scriptsize
\centering
\resizebox{\linewidth}{!}{
\begin{tabular}{lp{10cm}l}
\hline
Creator Description & \textbf{Mike Rankin} welcomes Sox in the Basement host \textbf{Chris Lanuti} to talk optimism on the South Side. The guys react to \textbf{Yoan Moncada}'s recent extension and later explore the potential for \textbf{Dylan Cease}, \textbf{Aaron Bummer} and well as\textbf{ Andrew Vaughn} in 2020. You can follow FutureSox ({\color{blue} @FutureSox}), Chris Lanuti ({\color{blue} @ChrisLanuti}) and Sox In The Basement ({\color{blue} @SoxInTheBasement}) on Twitter. \\\hline
     & \textbf{Generated Summaries} \\\hline
     
\textbf{\em bartcnn} & Hi. Hello and welcome to another episode of the Future Socks podcast. My name is \textbf{Mike Rankin}. I will be your host this time around and boy. What's going on? I appreciate you jumping in with us to talk some White Sox baseball on the future Sox podcast. You talked about socks in the basement for a couple of years now take me back to the Art, and what was the inspiration and then the growth across those two seasons? Well, the funny thing is that I had been doing podcasting for like a decade just as a hobby I used to do morning radio, but I never did anything in Chicago. I was in Champaign, Illinois and Reno, Nevada and Wheeling, West Virginia Pittsburgh and Southern California and Bakersfield, which is the armpit of America. It's like a hundred twenty two degrees in the shade and I'm from Chicago. And so when I came back here and you know started doing Thing that was out of radio. I still had the bug and I did this show called the broadcast basement and it was just a bunch of my friends used to be in radian. We'd mess around and somewhere along the line a couple years ago.
    \\\hline

\textbf{\em bartpodcasts} & Socks in the Basement" host \textbf{Chris Ludacris} joins \textbf{Mike Rankin} of the Future Sox Podcast to talk all things White Sox baseball, including the recent success of the team and what it means for the future of the franchise.   Chris is the host of the Socks In The Basement podcast, which is a weekly podcast covering the Chicago White Sox. You can follow them on Twitter {\color{blue} @socksinthebasepod} and check out the show at {\color{blue} socksinthesound.com.}   You can also follow the show on Twitter at {\color{blue} @SocksInTheBasetPod} and on Instagram at {\color{blue} SocksinTheBasepod.com\/socks\_in\_the\_baset}. The show is brought to you by the Chicago Sun-Times and Wrigley Field.   \\\hline

\textbf{\em category-aware-1epoch } & On this episode of the Future Sox Podcast, \textbf{Chris Ludacris} from Socks in the Basement joins \textbf{Mike Rankin} to talk about the White Sox's 2019 season and what we expect from the Sox in 2020. We also talk about how the Sox are getting better on the field and off the field, and how we expect the Sox to get better in 2020 as well. We hope you enjoy this episode, and if you do, please let us know what you thought of the show and what you would like to see in 2020 by leaving us a review on your favorite podcast platform. We love to hear your thoughts and feedback on the show, so feel free to reach out to us on Twitter at {\color{blue} @SocksInTheBasementPod} and we'd love to have you on the podcast in the future. \\\hline

\textbf{\em category-aware-2epochs } & \textbf{Chris Ludacris} of Socks in the Basement joins the show to talk about the White Sox offseason and the future of the team as well as his podcast Socks In The Basement and how it has grown to where it is today. \\\hline

\textbf{\em coarse2fine} & Socks in the Basement"" is a weekly podcast covering the Chicago White Sox baseball team through the eyes of Sox fans. Hosts are {\color{blue} @BBWhiteSox}, {\color{blue} @SoxOnTheBus}, and {\color{blue} @BBBlackSox}. This week we discuss the White Sox's bullpen, \textbf{Jose Abreu},\textbf{ Jose Ramirez}, \textbf{Dylan Cease}, and much more. You can follow us on Twitter {\color{blue} @SocksInTheBasementPod} and check out the extra content about the show at {\color{blue} SoxOnTheBaset.com}. \\\hline
\end{tabular}}
\caption{Example summaries generated by various models for a sports podcast episode (\url{ https://open.spotify.com/episode/5KmWK8Qh5lsWIb2sUuJp4r}).}
\label{tab:anecdotes}
\end{table*}

As an illustrative example, we present the summaries of the models as well as the creator description of one selected episode of a sports podcast in Table \ref{tab:anecdotes}. The creator description contains different named entities (in \textbf{bold}) as well as social media information and links (in {\color{blue} blue}). The {\em bartcnn} summary is verbose and long and captured the majority of the information from the beginning of the transcript, following the structure of the news summarizer. {\em bartpodcasts} and {\em category-aware-1epoch}, on the other hand, are similar to the creator description.
The {\em category-aware-2epochs} is the shortest compared to the other models, however, it includes the main theme and the majority of the necessary information. Finally, the {\em coarse2fine} summary does not include the names of the host or guest but it is the only one that contains additional named entities, similar to the creator description. 

NIST judges also assessed the quality of the summaries by answering eight yes/no questions about podcast summary attributes. A good summary will receive `yes' for all questions except Q6 (``Does the summary contain redundant information?''). With respect to Q6, Q7, and Q8, our {\em category-aware-2epoch} model generates the most coherent summaries with the least amount of redundant information compared to the baselines (Table \ref{tab:q1-q8_nist}) suggesting that the model may be learning aspects of style.

\begin{table*}[ht!]
\small
\centering
\resizebox{\linewidth}{!}{
\begin{tabular}{|p{2.3in}|>{\centering}p{0.5in}|>{\centering}p{0.6in}|>{\centering}p{0.5in}|>{\centering}p{0.5in}|p{0.5in}<{\centering}|}
\hline
\textbf{Attribute} & \textbf{\em bartcnn} & \textbf{\em bartpodcasts} & \textbf{\em category-aware-1epoch} & \textbf{\em category-aware-2epochs} & \textbf{\em coarse2fine} \\ \hline
Q1: Does the summary include names of  the main people (hosts, guests, characters) involved or mentioned in the podcast? & \textbf{72.6} & 51.4 & 64.2 & 55.3 & 35.2 \\ \hline
Q2: Does the summary give any additional information about the people mentioned (such as their job titles, biographies, personal background, etc)? & \textbf{48.0} & 34.6 & 40.2 & 40.8 & 22.9 \\ \hline
Q3: Does the summary include the main topic(s) of the podcast? & 69.3 & \textbf{78.2} & 76.5 & 77.7 & 60.3 \\ \hline
Q4: Does the summary tell you anything about the format of the podcast; e.g. whether it’s an interview, whether it’s a chat between friends, a monologue, etc? & 58.1 & 49.2 & \textbf{59.2} & 57.0 & 41.3 \\ \hline
Q5: Does the summary give you more context on the title of the podcast? & \textbf{67.6} & 65.4 & 65.4 & 65.9 & 57 \\ \hline
Q6: Does the summary contain redundant information? & 25.7 & 14.0 & 12.8 & \textbf{6.7} & 10.1 \\ \hline
Q7: Is the summary written in good English? & 34.6 & 86.6 & 86 & \textbf{88.0} & 79.9 \\ \hline
Q8: Are the start and end of the summary good sentence and paragraph start and end points? & 22.3 & 62 & 69.8 & \textbf{70.4} & 55.3 \\ \hline
\end{tabular}
}%
\caption{Average percentage of `Yes' judgments by human evaluators for each of the summary attributes.}
\label{tab:q1-q8_nist}
\end{table*} 

Overall, the automatic and manual evaluations of our models show that our {\em category-aware} models outperformed the baseline resulting in high quality summaries. The results of the submitted summaries show that our best model achieves an aggregate quality score (as defined by the task) of 1.58 in comparison to the creator descriptions and a baseline abstractive system which both score 1.49 (an improvement of 9\%) as assessed by human evaluators (Table \ref{tab:egfb}).

\section{Conclusion}
In this paper, we showcased two abstractive summarization models, one that is informed by the categories of the podcasts, and a hybrid model that uses an extractive step biased towards named entities. Experimental results show that our category-aware models generate better summaries than the baselines, but the model intended to bias summaries to named entity mentions underperforms. In future, we hope to investigate new and robust ways of biasing the summarizer to generate named entities in the summaries.

\bibliography{acl2020}
\bibliographystyle{acl_natbib}

\end{document}